\documentclass[conference,twoside]{IEEEtran}
\IEEEoverridecommandlockouts
\usepackage[a4paper,top=0.825in,bottom=0.825in,right=0.68in,left=0.68in]{geometry}
\usepackage{fancyhdr}

\usepackage[dvips]{graphicx}
\usepackage{eqparbox}
\usepackage{framed,multirow}
\usepackage{amssymb,dsfont}
\usepackage{latexsym}
\usepackage{url}
\usepackage[table]{xcolor}
\usepackage{color}
\usepackage[tight,footnotesize]{subfigure}
\usepackage{tabularx, tabulary, multirow, bigstrut}
\usepackage[american]{babel}
\usepackage[cmex10]{amsmath}
\usepackage{caption}
\usepackage[boxruled,vlined]{algorithm2e} 
\usepackage{cite}

\usepackage[draft]{hyperref} 
\hypersetup{
    colorlinks,
    linkcolor={red!50!black},
    citecolor={blue!50!black},
    urlcolor={blue!80!black}
}

\graphicspath{ {./figs/} }

\begin{document}
%
\title{Efficient Asymmetric Co-Tracking using Uncertainty Sampling}

\author{\IEEEauthorblockN{Kourosh Meshgi, Maryam Sadat Mirzaei, Shigeyuki Oba, Shin Ishii}
\IEEEauthorblockA{Graduate School of Informatics\\
Kyoto University\\
Sakyo-ward, Yoshida-honmachi, Kyoto 606--8501\\
Email: {meshgi-k}@sys.i.kyoto-u.ac.jp}
}

\maketitle
\renewcommand{\headrulewidth}{0pt}
\renewcommand{\footrulewidth}{0pt}
\thispagestyle{fancy}
\lfoot{978-1-4799-8996-6/15 \$31.00 \copyright 2017 IEEE}
\cfoot{}
\fancyhead[LO,RE]{{\small 2017 IEEE International Conference on Signal and Image Processing Applications (ICSIPA)}}

\begin{abstract}
Adaptive tracking-by-detection approaches are popular for tracking arbitrary objects. They treat the tracking problem as a classification task and use online learning techniques to update the object model. However, these approaches are heavily invested in the efficiency and effectiveness of their detectors. Evaluating a massive number of samples for each frame (e.g., obtained by a sliding window) forces the detector to trade the accuracy in favor of speed. Furthermore, misclassification of borderline samples in the detector introduce accumulating errors in tracking. In this study, we propose a co-tracking based on the efficient cooperation of two detectors: a rapid adaptive exemplar-based detector and another more sophisticated but slower detector with a long-term memory. The sampling labeling and co-learning of the detectors are conducted by an uncertainty sampling unit, which improves the speed and accuracy of the system. We also introduce a budgeting mechanism which prevents the unbounded growth in the number of examples in the first detector to maintain its rapid response. Experiments demonstrate the efficiency and effectiveness of the proposed tracker against its baselines and its superior performance against state-of-the-art trackers on various benchmark videos.
\end{abstract}

%
\IEEEpeerreviewmaketitle

\section{Introduction}
Nowadays, visual tracking is an inseparable component for high-level visual tasks such as human-computer interface, human behavior analysis, smart appliances, virtual/augmented reality and surveillance. When applied to video sequences in real-life situations, trackers should cope with challenging appearance changes due to illumination variations, motion blur, non-rigid deformations, rotations, mobile imaging platforms and occlusions \cite{wu2015object}. 
While some successful generative trackers \cite{adam2006robust,zhang2012real,oron2015locally,taalimi2015online} models the target object, they ignore the information hidden in the background. On the other hand discriminative trackers 
\cite{stalder2009beyond,babenko2009visual,grabner2010tracking,hare2011struck,dinh2011context,kalal2012tracking,henriques2012exploiting}
pose the tracking problem as a classification task. In these models instead of trying to build a complex model of the object, the algorithms seek a decision boundary that best separates the target and background. This re-formulation undermine the inherent issues of generative models like background clutter and model over-simplification \cite{tang2007co}.

\begin{figure}[!t]
\centering
\includegraphics[width=0.8\linewidth]{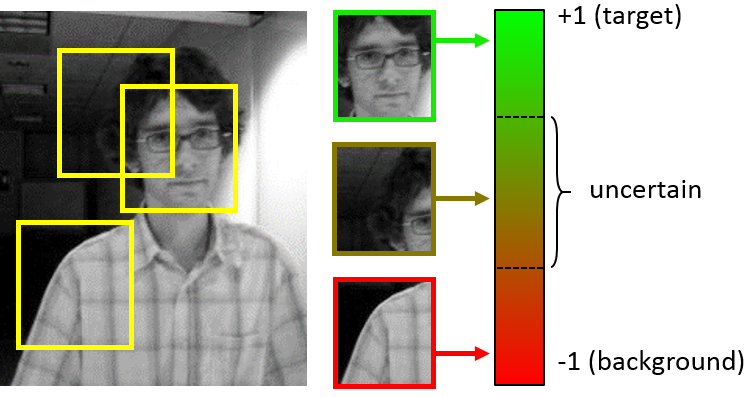}
\caption{Labeling dilemma in tracking-by-detection. The classifier of the tracker can label the target and the background  only when it has high confidence. Labeling the uncertain patches may result in label noise. }
\label{fig:uncertain}
\vspace{-0.5 cm}
\end{figure}

There are many tracking-by-detection visual trackers, which heavily rely on their detector to handle different tracking challenges \cite{hare2011struck}, such as rotations and scale changes. Such schemes treat tracking as a binary classification problem, which separates the object from its local background using a classifier, in which a discriminative classifier is trained with the samples obtained from the tracking sequence, and their performances are affected by their sampling policy. Most trackers only utilize one positive sample, i.e., the tracking result in the current frame \cite{grabner2006real}. If the tracked location is not accurate, the classifier will be updated with the contaminated appearance of the target, leading to a drift over time. To alleviate this problem, multiple samples in the proximity of the estimated target can be used to train the tracker \cite{babenko2009visual,hare2011struck}. However, such algorithms are heavily invested in the efficiency and effectiveness of their detectors. Evaluating a massive number of samples for each input frame forces the detector to trade the accuracy in favor of speed to meet the real-time processing requirements. While some trackers aim to enhance the detectors' speed while preserving their accuracy using statistical properties of images (e.g., \cite{henriques2012exploiting}), generally achieving an adjustable balance between speed and accuracy is desired. Furthermore, dealing with rotations and scale changes challenges such mechanisms. Additionally, misclassification of the borderline input samples in the detector (Figure \ref{fig:uncertain}) may introduce accumulating errors in the tracker, degrading its performance significantly \cite{babenko2009visual}. Furthermore, the growth of sample repository in online learning schemes degrades the speed. If not handled properly, the tracker cannot perform long-term tracking \cite{hare2011struck}.

\begin{figure*}
\centering
\includegraphics[width=1\linewidth]{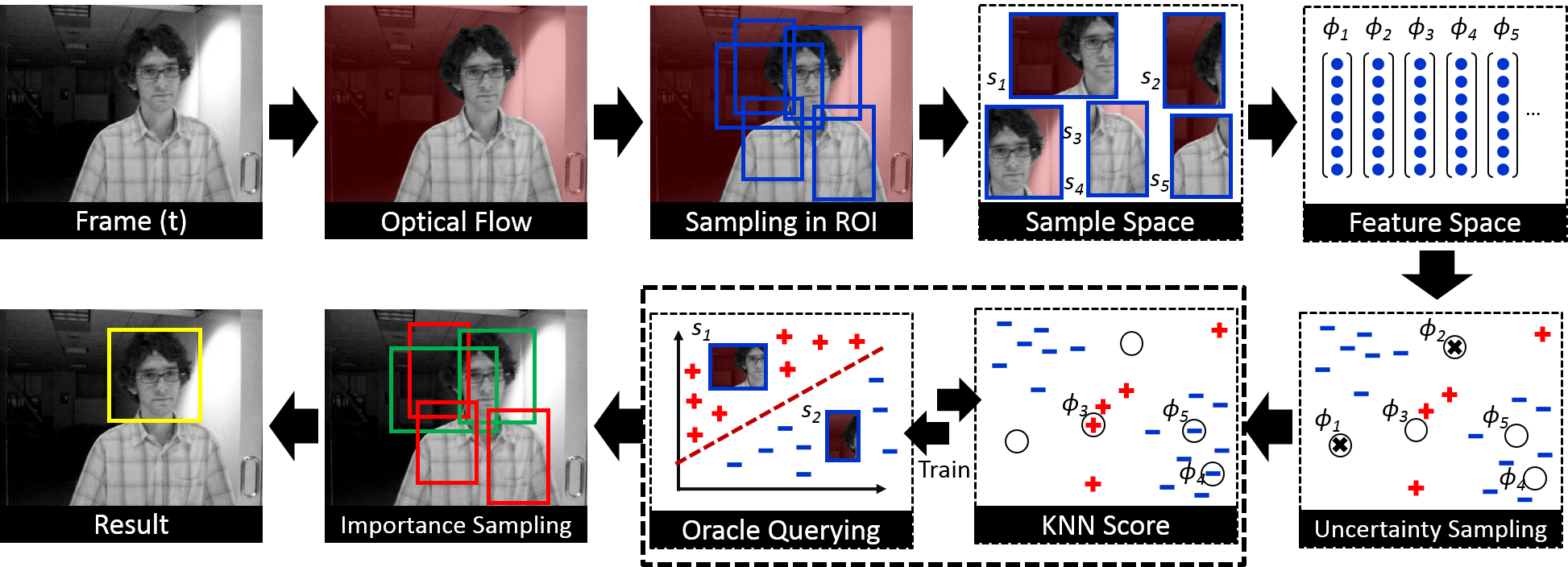}
\caption{The flow of the proposed tracker, UST. After sampling from the region-of-interest determined by optical flow, the samples are mapped onto a low dimension space, and collaboratively labeled by the two classifiers. This collaboration is organized by uncertainty sampling unit. The labeled are used to localize the target through a weighted averaging, and the two classifiers provide training data for each other.}
\label{fig:schematic}
\vspace{-0.7 cm}
\end{figure*}

In this study, we propose an efficient co-tracking framework in which an active learning unit orchestrate the information exchange. It consists of a rapid detector with short-term memory, and an uncertainty sampling switcher that query the label of the most uncertain samples of the first detector from an accurate detector with a long-term memory (called the \textit{``oracle''}). An importance sampling scheme combines the results of the two trackers and handles the scale variation of the target. An exemplar-based detector is employed as the rapid detector and we introduce a budgeting mechanism to prevent the unbounded growth in the number of examples in this detector to maintain its rapid response. In summary we \textit{(i)} employed active learning in co-tracking framework that leads to increasing the speed and generalization power of the tracker, \textit{(ii)} actively control the memory of tracker by balancing between short- and long-term memories and \textit{(iii)} introduced an intuitive budgeting method for a nearest neighbor classifier. 

The difference between the proposed co-tracking framework and that of Tang et al. \cite{tang2007co} is four-fold: \textit{(i)} the classifiers do not exchange all the data they have problems in labeling, instead, the most informative samples are selected by uncertainty sampling, and exchanged. \textit{(ii)} the update rate of classifiers is different to realize a short and long-term memory mixture, \textit{(iii)} the samples that are labeled for the target localization can be re-used for training and the need for an extra round of sampling and labeling is revoked, \textit{(iv)} since in the proposed asymmetric co-tracking, one of the classifiers scaffolds the other one instead of participating in every labeling process, a more sophisticated classifier with higher computational complexity can be used.

\section{Related Work}
\label{sect:related}
Many discriminative models have been adopted in object tracking, where a binary classifier is learned online to separate the target from the background. Numerous supervised or semi-supervised classifiers have been employed for object tracking, such as SVM \cite{avidan2004support}, structured output SVM \cite{hare2011struck}, boosting \cite{grabner2006real}, semi-boosting \cite{grabner2008semi}, and online multi-instance boosting \cite{babenko2009visual}. They follow different approaches in tackling foreground-background separation like incorporating a trained SVM into an optical flow tracker \cite{avidan2004support}, using an ensemble of online learned weak classifiers to decides whether a pixel belongs to the target region or background \cite{avidan2007ensemble}, or utilizing online boosting to select discriminative features for separation of target and background \cite{grabner2006real}. Multiple instance learning (MIL) tracker put all of the ambiguous positive and negative samples into bags to learn a discriminative model \cite{babenko2009visual}. In another stream of studies, the most discriminative feature combination in learned online to build a confidence map for foreground detection \cite{collins2005online}. Combining multiple supervised and semi-supervised classifiers \cite{stalder2009beyond}, and governing a learning method using positive and negative constraints \cite{kalal2012tracking} are some other of the most successful discriminative approaches for tracking. Such methods are specifically designed to resolve the label noise problem, in which the classifier get confused by even the smallest mistakes in the labeling process. Since the classifier is using a self-learning loop, such mistakes can accumulate over time and cause the tracker to drift. One of the solutions to this problem is co-tracking \cite{tang2007co}, in which the self-learning loop is broken and labeling is done collaboratively. 

Without model update schemes, trackers accumulate error during run-time (drift) and typically fail if the object disappears temporarily. To address this issue, some online appearance update models have been proposed, e.g., incremental subspace update scheme and adaptive sparse representation update \cite{bao2012real}. In the case of tracking-by-detection approaches, this essentially means that the classifier should be re-trained with the relevant samples. In such algorithms, after selecting the samples and labeling them, they are used to update the classifier. The classifier aims for labeling data, while the tracker attempts to localize the target and these two objectives are sometimes contradicting (e.g., in the presence of target-like distractors). Increasing the accuracy of the detector and using unlabeled samples is a typical approach to this problem (e.g., \cite{grabner2008semi,babenko2009visual}), while trackers such as STRUCK \cite{hare2011struck} couple these two objectives in a joint learning framework. Another instance was presented in \cite{kalal2012tracking} where the recent samples are added to the classifier only if their classifier label is different from the label of constrained classifiers that monitors the performance of the tracker. Combining short and long-term memory to deal with rapid changing targets, occlusions, and environmental change in tracking is another research avenue for model updating \cite{hong2015multi}, however, the proposed schemes are hardly integrated into general trackers.

Active learning techniques build upon such discriminative models and try to improve their convergence speed and generalization power. For instance, Fisher information criterion evaluates the uncertainty of classification model in the MIL tracker \cite{babenko2009visual} to perform active feature selection \cite{zhang2013robust}. Reducing the number of necessary labeled samples \cite{lampert2009active}, unified sample learning and feature selection procedure \cite{li2015active} and reducing the sampling bias by controlling the variance \cite{beygelzimer2009importance} are some of the improvements that active learning provides for the discriminative trackers. Active learner selects the samples that don't know how to label. Uncertainty sampling \cite{lewis1994sequential}, as one of the most popular forms of active learning, try to query the sample that minimizes a utility function from the oracle. The utility function can be classification confidence \cite{lewis1994heterogeneous}, margin \cite{scheffer2001active}, and entropy \cite{settles2008analysis}. Uncertainty sampling in this sense, optimize the query selection with respect to the utility function. However, this approach is a form of Gibbs sampling and requires probabilistic learning models, it should be treated differently for non-probabilistic classifiers. In this regard, decision trees \cite{lewis1994heterogeneous} and nearest neighbors \cite{lindenbaum2004selective} are used with uncertainty sampling with the class label obtained by voting, and for SVMs \cite{tong2002support} the proximity to the decision boundary is considered as the utility function. 

\section{Proposed Method}
\label{sect:proposed}

\subsection{Architecture of the System}
\label{sect:arch}
The proposed tracker (Figure \ref{fig:schematic}) consists of two classifiers $\theta^{(1)}_t$ and $\theta^{(2)}_t$, who exchange their information using an uncertainty sampling scheme. The samples are obtained from the region-of-interest (ROI) defined by optical flow with Gaussian probability. Then, the samples are labeled by a collaborative effort of the classifiers. Later, these samples and their labels are employed to update the classifiers. The first classifier ($\theta^{(1)}_t$) is a short-memory highly adaptive exemplar-based classifier that is updated (with respect to a memory budget) with the most informative samples in each tracking episode. The second classifier, the oracle ($\theta^{(2)}_t$), is a long-term memory tracker that is updated with all the samples at fixed intervals to grant the robustness against occlusions and temporal target changes to the tracker.

\textbf{Sampling:} In each frame $F_t , t \in \{1, \ldots , T\}$, a set of $n$ random samples $\mathbf{p}^j_t$ is generated. These samples are selected from a region-of-interest $\mathcal{R}_t$ determined by optical flow \cite{brox2009large}. To handle still objects, the last known target area is added to the ROI 
New samples are selected from ROI, based on a Gaussian distribution centered on the last target position $\mathcal{N}(\mathbf{p}_{t-1},\Sigma_{search})$
. In each frame, an additional $n'$ samples are selected uniformly from distant locations of the frame (distance larger than $3\times\Sigma_{search}$), to sample global background, and are automatically labeled as the background. 
This sampling scheme empowers the foreground-background separability in short term by exploiting the locality of the target \cite{hare2011struck}. Furthermore, it enables the tracker to handle in-scene distractors (e.g., the moving non-target objects in the ROI) and potential occluders using the global sampling for negative exemplars. 

\begin{algorithm}[!t]
\DontPrintSemicolon
\SetKwInOut{Input}{input}\SetKwInOut{Output}{output}

\Input{Target position in last frame $\mathbf{p}_{t-1}$}
\Output{Target position in current frame $\mathbf{p}_{t}$}
\BlankLine

\emph{ROI $\mathcal{R}_t \leftarrow OpticalFlow (F_{t-1},F_t) \cup Area(\mathbf{p}_{t-1})$}\;
\For{$j \leftarrow 1$ \KwTo $n$ }
{
\emph{Make a sample $\mathbf{p}_t^j \sim \mathcal{N}(\mathbf{p}_{t-1},\Sigma_{search})$}\;
\uIf(){ $\mathbf{p}_t^j \in \mathcal{R}_t$}
	{
	\emph{Calculate $s_t^j \leftarrow h \big( \mathbf{x}_t^{\mathbf{p}_t^j} | \theta_t^{(1)} \big)$} (eq\eqref{eq:score})\;
	\emph{Distinguish uncertain samples $\mathcal{U}_t$} (eq\eqref{eq:uncertain})\;
	\uIf($\theta_t^{(1)}$ is uncertain){$\mathbf{p}_t^j \in \mathcal{U}_t$}
		{
		\emph{Query $\theta_t^{(2)}$: $\ell_t^j \leftarrow Sign \Big( h \big( \mathbf{x}_t^{\mathbf{p}_t^j} | \theta_t^{(2)} \big) \Big)$}\;		
		}
	\Else()
		{
		\emph{Label using $\theta_t^{(1)}$: $\ell_t^j \leftarrow Sign( s_t^j )$}\;
		}
	}
\Else(Sample not in ROI)
	{
	\emph{$\ell^j_t \leftarrow -1$ (local background)}\;
	}
\emph{$\mathcal{D}_t \leftarrow \mathcal{D}_t \cup \langle \mathbf{x}_t^{\mathbf{p}_t^j} , \ell^j_t  \rangle$}\;
}
\emph{Create $n'$ global background samples and add to $\mathcal{D}_t$}\;
\emph{Update $\theta_t^{(2)}$ with $\mathcal{D}_{t-\Delta,..,t}$ every $\Delta$ frames} (eq\eqref{eq:update2})\;
\uIf(){$\sum_{j=1}^n \mathds{1}(\ell^j_t > 0) > \tau_p$ and $\sum_{j=1}^n \pi_t^j > \tau_a$}
		{
		\emph{Approximate target state $\mathbf{\hat{p}}_t$} (eq\eqref{eq:approx}) \;
		\emph{Update $\theta_t^{(1)}$ with $\mathcal{U}_t$} (eq\eqref{eq:update1})\;
		}
	\Else(target occluded)
		{
		\emph{$\mathbf{\hat{p}}_t \leftarrow \mathbf{p}_{t-1}$}\;		
		}
\BlankLine

\caption{Uncertainty Sampling co-Tracker (UST)}
\label{alg:ust}
\end{algorithm}

\textbf{Labeling:} In the proposed asymmetric co-tracking framework, one classifier attempts to label the sample, and it queries the label from the other classifier if a certain condition is met. This is in contrast with using a linear combination of both classifiers based on their general classification accuracy, as adopted in \cite{tang2007co}. The proposed tracker can decide for each sample based on the classifier confidence, i.e., for sample $\mathbf{p}_t^j$ we define a score $s^j_t$
\begin{equation}
s^j_t = h \big( \mathbf{x}_t^{\mathbf{p}_t^j} | \theta_t^{(1)} \big)
\label{eq:score}
\end{equation}
that reflects the classification score for the image patch $\mathbf{x}_t^{\mathbf{p}_t^j}$, with values closer to +1 as possible targets and values closer to -1 as  background. Based on uncertainty sampling (elaborated in section \ref{sect:uncertainty}), the samples for which the classification score is more uncertain (i.e., $s^j_t \rightarrow 0$), contains more information for the classifier if they are labeled by the other classifier (i.e., the oracle). Therefore, the scores of all samples are sorted, and $m$ samples with the closest values to 0 are selected to be queried from $\theta_t^{(2)}$. To handle the situations for which the number of highly uncertain samples are more then $m$, a range of scores are defined by lower and higher thresholds ($\tau_l$ and $\tau_u$) and all the samples in this range are considered highly uncertain.
\begin{equation}
\mathcal{U}_t = \{ \mathbf{p}^i_t | \tau_l < s_t^i < \tau_u \, \mathrm{or} \, \Bigm\lvert \{\exists j \neq i | s^j_t \leq s^i_t \} \Bigm\lvert <m \}
\label{eq:uncertain}
\end{equation}
in which $\mathcal{U}_t$ is the list of uncertain samples. The label of the samples $\ell^j_t \in \mathcal{L}_t$ are then determined by
\begin{align}
\ell^j_t &=
  \begin{cases}
   sign \Big( h \big( \mathbf{x}_t^{\mathbf{p}_t^j} | \theta_t^{(1)} \big) \Big)        & , \mathbf{p}_t^j \in \mathcal{U}_t\\
   sign \Big( h \big( \mathbf{x}_t^{\mathbf{p}_t^j} | \theta_t^{(2)} \big) \Big)        & , \mathbf{p}_t^j \notin \mathcal{U}_t\\
  \end{cases}
\label{eq:label}
\end{align}
and all image patches $\mathbf{x}_t^{\mathbf{p}_t^j}$ and labels $\ell^j_t$ are stored in $\mathcal{D}_t$.

\textbf{Localizing:} To determine the state of the target $\hat{\mathbf{p}_t}$, we follow the importance sampling mechanism originally employed by particle filter trackers, 
\begin{equation}
\hat{\mathbf{p}_t} = \frac{\sum_{j=1}^n \pi_t^j \mathbf{p}_t^j}{\sum_{j=1}^1 \pi_t^j}.
\label{eq:approx}
\end{equation}
where $\pi^j_t = s^j_t \mathds{1}(\ell^j_t > 0)$ and $\mathds{1}(.)$ is the indicator function, 1 if true, zero otherwise.
This mechanism approximate the state of the target, based on the effect of positive samples, in which samples with higher scores gravitates the final results more toward themselves. Upon the events such as massive occlusion or target loss, this sampling mechanism degenerates \cite{bao2012real}. In such cases, the number of positive samples and their corresponding weights shrinks significantly, and the importance sampling is prone to outliers, distractors and occluded patches. To address this issue, two thresholds ($\tau_p$ and $\tau_a$) are set on the number and average scores of positive samples. If either of thresholds are not exceeded, the target is deemed occluded to avoid tracker degeneracy. 

\begin{figure}[!t]
\centering
\includegraphics[width=1\linewidth]{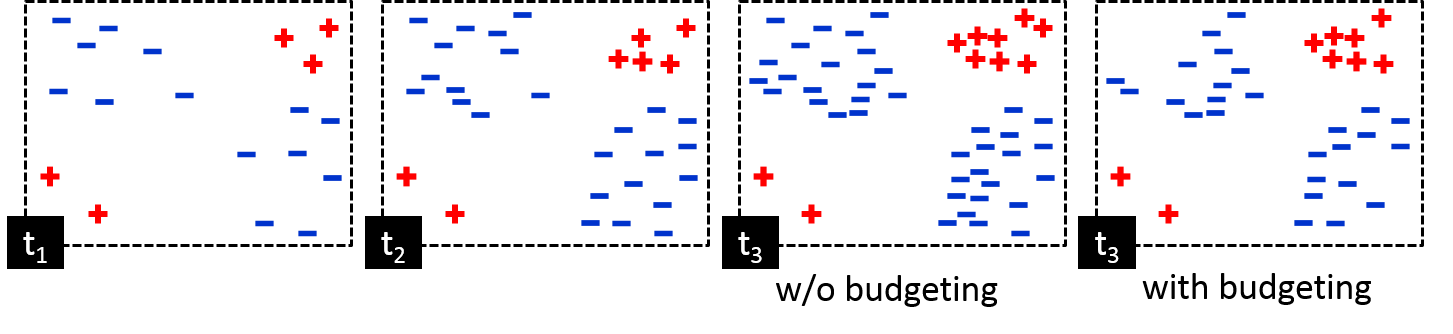}
\caption{An ever growing example-space of a KNN and the budgeting mechanism to discard the most futile samples. Without budgeting, some samples are added/kept by the classifier that has no effect on the discrimination border. }
\label{fig:knn_budget}
\vspace{-0.7 cm}
\end{figure}

\textbf{Updating:} 
In the proposed tracker, the first classifier, $\theta_t^{(1)}$ is updated using the samples that it queried from $\theta_t^{(2)}$, i.e, with the those it was uncertain about and had oracle label them ($\mathcal{U}_t$). This uncertainty is  either due to the model shortcomings of the classifier (e.g., simple observation model) or because of intrinsic ambiguity of the sample. Using a sophisticated classifier as the oracle alleviate these issues, and by providing the label back to the first classifier, it scaffolds it for better classification in similar circumstances, potentially improving the speed of future sample evaluation as well as the generalization.
\begin{equation}
\theta_{t+1}^{(1)} = \psi( \theta_t^{(1)}, \mathcal{U}_t, \mathcal{L}_t )
\label{eq:update1}
\end{equation}
in which $\psi(.)$ is the update function (r.t. \ref{sect:budget}).
On the other hand, to realize a dual-memory scheme to handle temporal target changes and occlusions, the oracle is equipped with a non-volatile memory and updated less frequently (every $\Delta$ frames) with all the data sampled during this period,
\begin{align}
\theta^{(2)}_{t+1} = 
    \begin{cases}
    u(\theta^{(2)}_t, \mathcal{D}_{t-\Delta,..,t})               &, \mathrm{if} \; t \neq k\Delta \\
    \theta^{(2)}_{t}                                             &, \mathrm{if} \; t = k\Delta
    \end{cases}
\label{eq:update2}
\end{align}
in which $u(.)$ is a classifier re-train function.
In summary, the first detector rapidly adapts to the target in order to estimate the target location considering its recent changes. The second detector, however, cares for best object detection performance, has a long-term memory and is robust to noise and temporal occlusions.  The co-training of these two detectors balances the desired level of speed and accuracy for the tracker. This differs from the dual memory scheme of MUSTer \cite{hong2015multi}, in which the short-term memory is based on the information obtained from the current frame, and the long-term memory that has exponential forgetting curve and get updated only when no occlusion is detected. 

\subsection{Realization}
\label{sect:realize}
In this study, the short-term memory classifier is implemented using a k-nearest neighbor classifier in which all the samples have a short lifetime to realize the budgeting mechanism. The histogram of colors and bag of visual words (of SIFT) forms the feature vector of every patch $\mathbf{x}^{\mathbf{p}^j_t}_t$, and its dimensionality is reduced to 20 using PCA. The KD-tree-based KNN classifier, its budgeted memory, the lazy update behavior of KNNs, and the reduced dimensionality of the feature space render the KNN suitable for real-time tracking. The oracle in this study is a part-based detector \cite{felzenszwalb2010object}. The features, part-base detector dictionary, and the parameters ($k$ of KNN, thresholds $\tau_l,\tau_u,\tau_a,\tau_p$, number of samples $n,n'$, search radius $\Sigma_{search}$, and update latency $\Delta$) are trained/tuned via a cross-validation approach.

\begin{figure}[!t]
\centering
\includegraphics[width=1\linewidth]{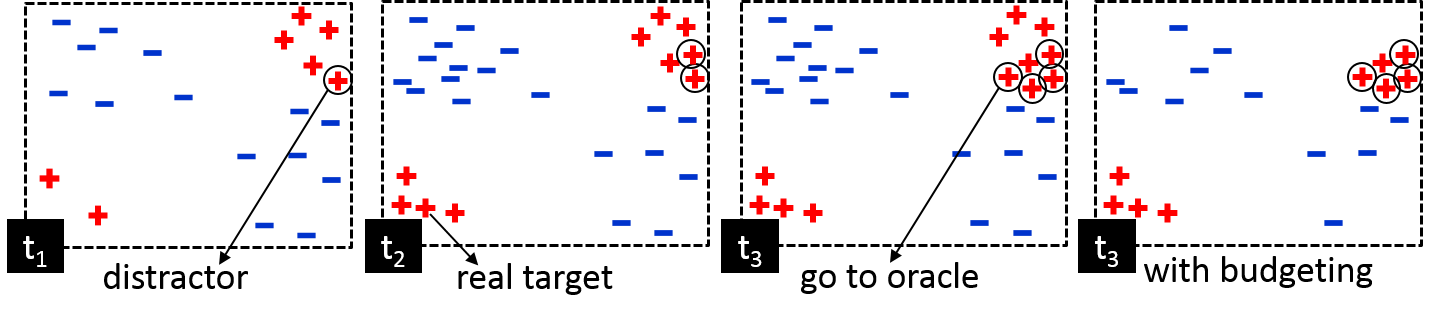}
\caption{KNN local minima and the budgeting method that help solving it. The budgeting prevent the distractor to reinforce itself by influencing the neighbors and spread throughout the sample space.}
\label{fig:knn_minima}
\vspace{-0.7 cm}
\end{figure}

\subsection{Budgeting} 
\label{sect:budget}
Online learning of discriminative trackers has its own challenges. The sample size of most of the adaptive classifiers is constantly growing, making them slower by the time. For a KNN classifier, even with a robust KD-tree architecture, the computation cost rapidly increase over time. This is similar to the curse of kernelization, in which the number of support vectors increases with the amount of training data in kernel classifiers. To allow for real-time operation, there is a need to control the number of support vectors. Recently, approaches have been proposed for online learning of classification SVMs on a fixed budget, meaning that the number of support vectors is constrained to remain within a specified limit as it is employed in \cite{hare2011struck}). 

Reducing the dataset for KNN classification has been studied in the literature (e.g., the condensed nearest neighbor \cite{angiulli2005fast}), yet it is not suitable for tracking in which the distribution of target and background is non-stationary, and there is a need to keep/remove the samples based on the temporal properties of tracking task. We propose a simple accounting method for a sample $\mathbf{x}$ with the nearest neighborhood $\eta(\mathbf{x})$ as KNN budgeting rules:
\begin{enumerate}
\item Discard the new sample $\mathbf{x}$ for which all $\eta(\mathbf{x})$ have similar labels (\textit{absorbed});
\item Attach a timer $\alpha$ to each new $\mathbf{x}$ that counts down upon processing each new frame. If the timer goes off ($\alpha \rightarrow 0$), the sample is flagged;
\item Mark the sample for which all neighbors have opposite labels as \textit{outlier};
\item For each added sample $\mathbf{x}$, increment the timer of all $\eta(\mathbf{x})$ whose labels differ with the new sample;
\item For each flagged sample if all of $\eta(\mathbf{x})$ has similar labels to $\mathbf{x}$ it is discarded (\textit{absorbed}), if $\mathbf{x}$ is an outlier and none of $\eta(\mathbf{x})$ has the same label, it should also be discarded. The remaining flagged samples are called \textit{prototypes}.
\end{enumerate}
This scheme tends to discard the most futile samples from the sample pool while preserving the recent or essential ones. Figure \ref{fig:knn_budget} depicts the sample 2D feature space of a KNN classifier and demonstrate how this budgeting mechanism preserve the classification power while reducing the number of samples. 
This budgeting scheme serves as a forgetting mechanism for the KNN classifier as well. If a distractor is very similar to the target in the feature space, the samples obtained from it will be labeled positive and added again to the KNN classifier, reinforcing the classifier false belief about the label of such samples. Such cases act as local minima for feature space in KNN classifier and can only be resolved if the oracle investigates and disprove them. To rescue the KNN classifier from its local minima, it requires a forgetting mechanism such as the proposed budgeting scheme. This concept is illustrated in Figure \ref{fig:knn_minima}.


\subsection{Uncertainty Sampling} 
\label{sect:uncertainty}
When using a probabilistic model for binary classification, target/non-target in the case of tracking-by-detection, uncertainty sampling simply queries the instance whose posterior probability of being is halfway between positive values and negative values \cite{lewis1994sequential}. Uncertainty sampling strategies may also be employed with non-probabilistic
algorithms like memory-based classifiers \cite{lindenbaum2004selective}. Inspired by these studies, we calculated the score of a sample as the KNN classifier confidence score, i.e., we allow the $k$ nearest neighbors to vote on the class label of $\mathbf{p}_t^i$, and the sum of these votes representing the score. 

As mentioned earlier, UST queries the $m$ most uncertain samples (having the closest scores to 0) from the oracle. To handle a large number of uncertain samples, we decided to query all the samples in the range of $(\tau_l,\tau_p)$ from the oracle. This backup mechanism along with forgetting mechanism realized by the budgeting helps the KNN detector to escape from its local minima induced by feature-space similar objects and partial occlusions (Figure \ref{fig:knn_minima}).

\section{Evaluation}
\label{sect:eval}
This section reports on a set of quantitative experiments comparing the UST with relevant algorithms. 
To evaluate the performance of the proposed tracker, the experiments is conducted on 100 challenging video sequences from \cite{wu2015object}. These sequences include many of the visual tracking challenges such as scale variation, fast motion and motion blur, illumination variations, in-plane and out-of-plane rotations, low resolution and shear problem, background clutter and various types of occlusion. The performance of the tracker is measured by the area under the surface of its success plot. A tracker in time $t$ succeed to track the object if its response $\hat{\mathbf{p}_t}$ overlaps with the ground truth $\mathbf{p}_t^*$ more than a threshold $\tau_o$. Success plot, graphs the success of the tracker against different values of the threshold $\tau_o$ and its AUC is
\begin{equation}
AUC = \frac{1}{T} \int_0^1 \sum_{t=1}^T \mathds{1} \left[ \frac{ | \hat{\mathbf{p}_t} \cap {\mathbf{p}_t^*} | }{ | \hat{\mathbf{p}_t} \cup {\mathbf{p}_t^*} | } > \tau_o \right] d_{\tau_o}
\end{equation}
where $\tau$ is the length of sequence, $|.|$ denotes the area of the region, $\cap$ and $\cup$ stands for intersection and union of the regions respectively. Since UST has non-deterministic sampling parts, we run it 5 times and the average of the results is reported.

\subsection{Comparison with Baseline}
\begin{figure}
\centering
\includegraphics[width=0.95\linewidth]{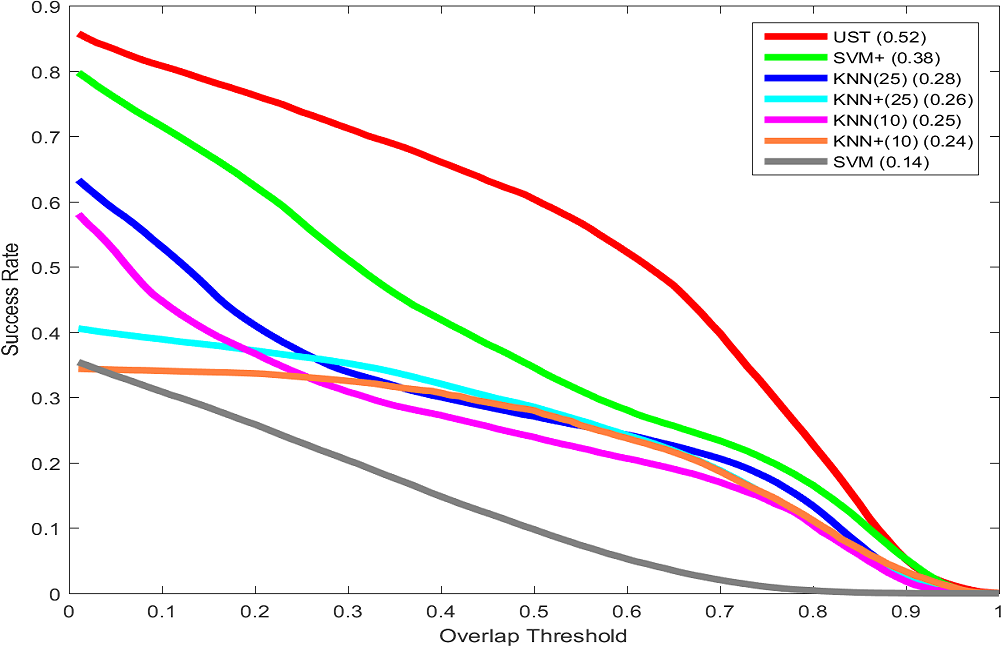}
\caption{Quantitative comparison of UST with its baselines}
\label{fig:cmp_base_suc}
\vspace{-0.7 cm}
\end{figure}

This experiment strives to demonstrate the advantages of the proposed tracker in comparison to the trackers that consists of either of its detectors in isolation. To this end, we construct several trackers from the components of this tracker to serve as the baselines for this experiment. In all of these trackers, the ROI detection and input sampling are similar to the UST tracker. KNN(10) and KNN(25) trackers utilize only the feature-based nearest neighbor detector for the tracking with the neighborhood size of $k=10$ and $k=25$ respectively. KNN+(10) and KNN+(25) trackers additionally incorporate the proposed budgeting mechanism into their detector. SVM tracker employs the oracle to track the target whereas SVM+ include the classifier update in its framework.
Figure \ref{fig:cmp_base_suc} compares the performance of these baseline tracker with that of UST, and demonstrates that the uncertainty sampling data exchange, effectively connect the classifiers to construct a robust and efficient tracker.

\begin{figure}[!t]
\centering
\includegraphics[width=0.95\linewidth]{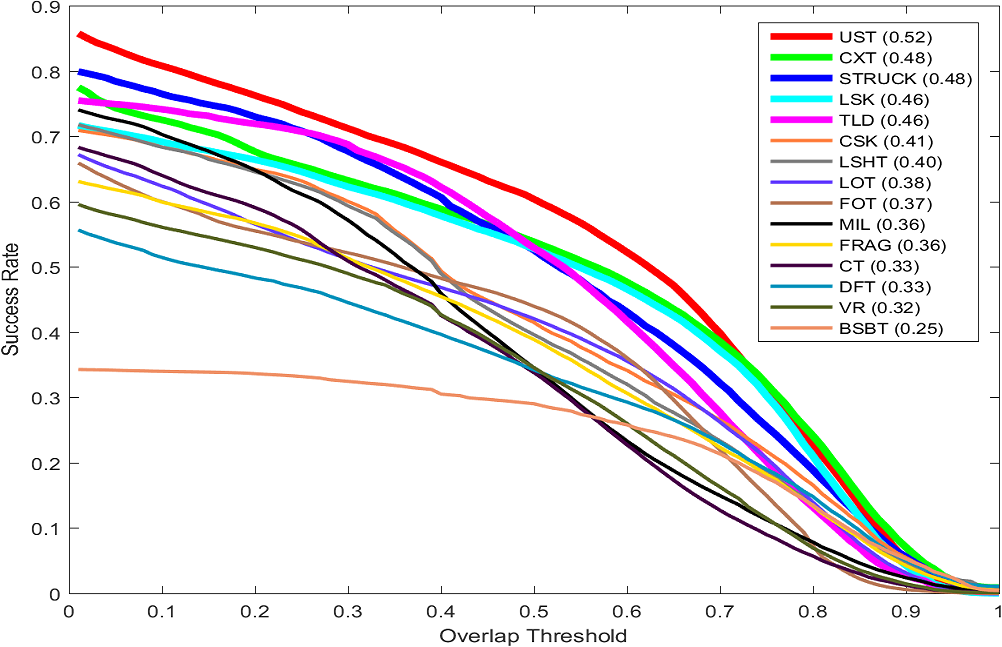}
\caption{Quantitative comparison of the proposed tracker, UST, with the state-of-the-art trackers using success plot} 
\label{fig:eval_succ_challenges}
\vspace{-0.7 cm}
\end{figure}

\subsection{Comparison with state-of-the-art}
To establish a fair comparison with the state-of-the-art, we select some of the most popular discriminative trackers based on \cite{wu2015object} and perform a benchmark on the whole videos of the dataset, along with partial subsets of the dataset with a distinguishing attribute to evaluate the tracker performance under different situations. These trackers are BSBT \cite{stalder2009beyond}, CSK \cite{henriques2012exploiting}, CT \cite{zhang2012real}, CXT \cite{dinh2011context}, 
DFT \cite{sevilla2012distribution}, FOT \cite{matas2011robustifying}, FRAG \cite{adam2006robust}, LOT \cite{oron2015locally}, LSHT \cite{he2013visual}, LSK \cite{liu2011robust}, MIL \cite{babenko2009visual}, SBT \cite{grabner2010tracking}, STRUCK \cite{hare2011struck}, TLD \cite{kalal2012tracking}, and VR \cite{collins2005online}. 

\begin{figure}[!t]
\centering
\includegraphics[width= 0.32\linewidth]{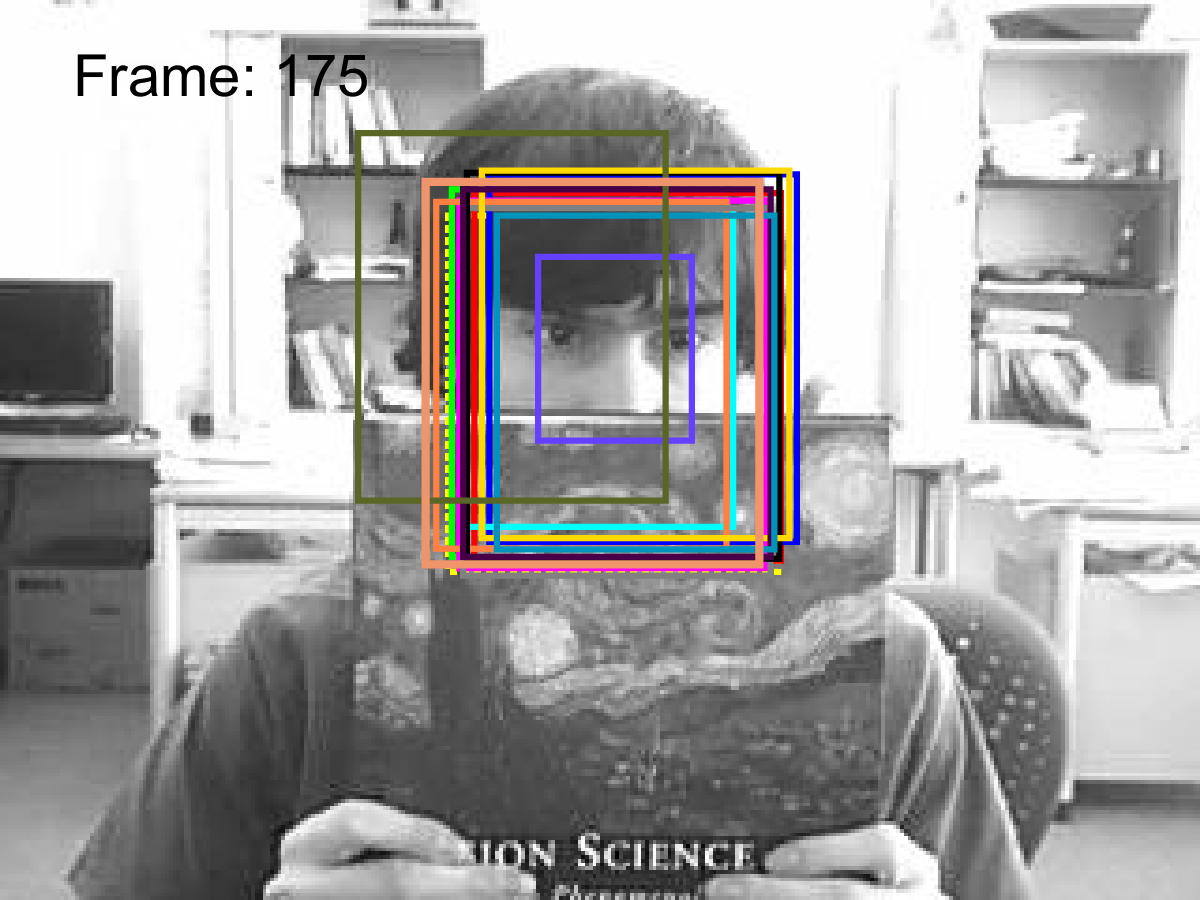}
\includegraphics[width= 0.32\linewidth]{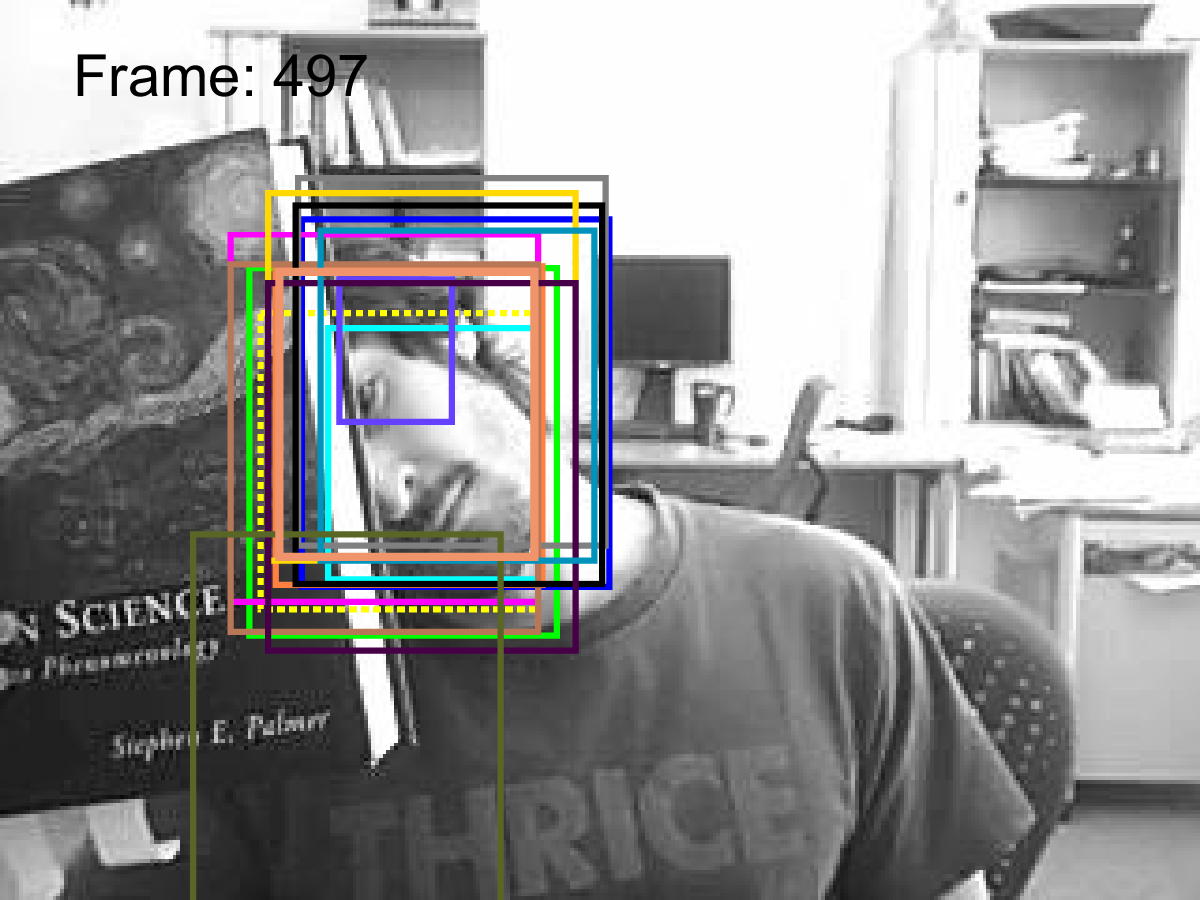}
\includegraphics[width= 0.32\linewidth]{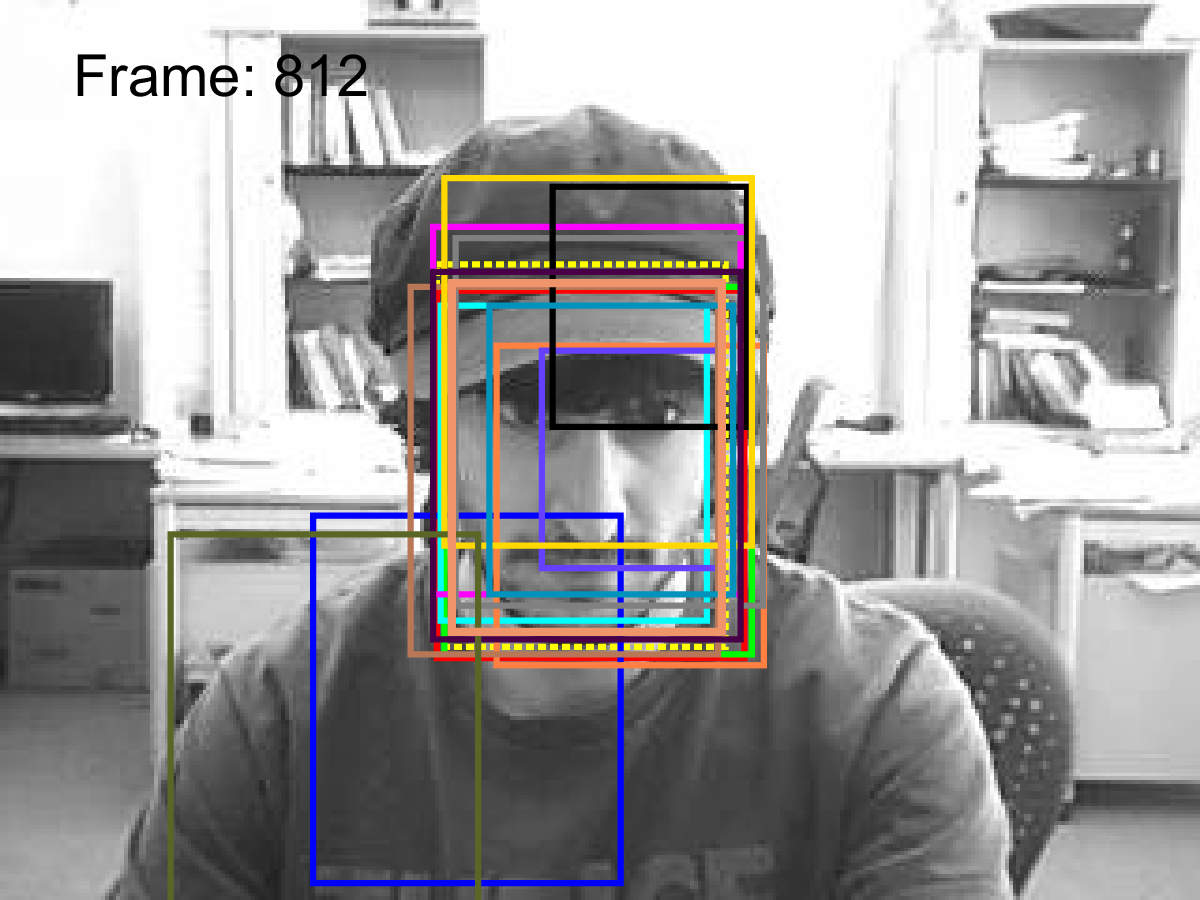}
\includegraphics[width= 0.32\linewidth]{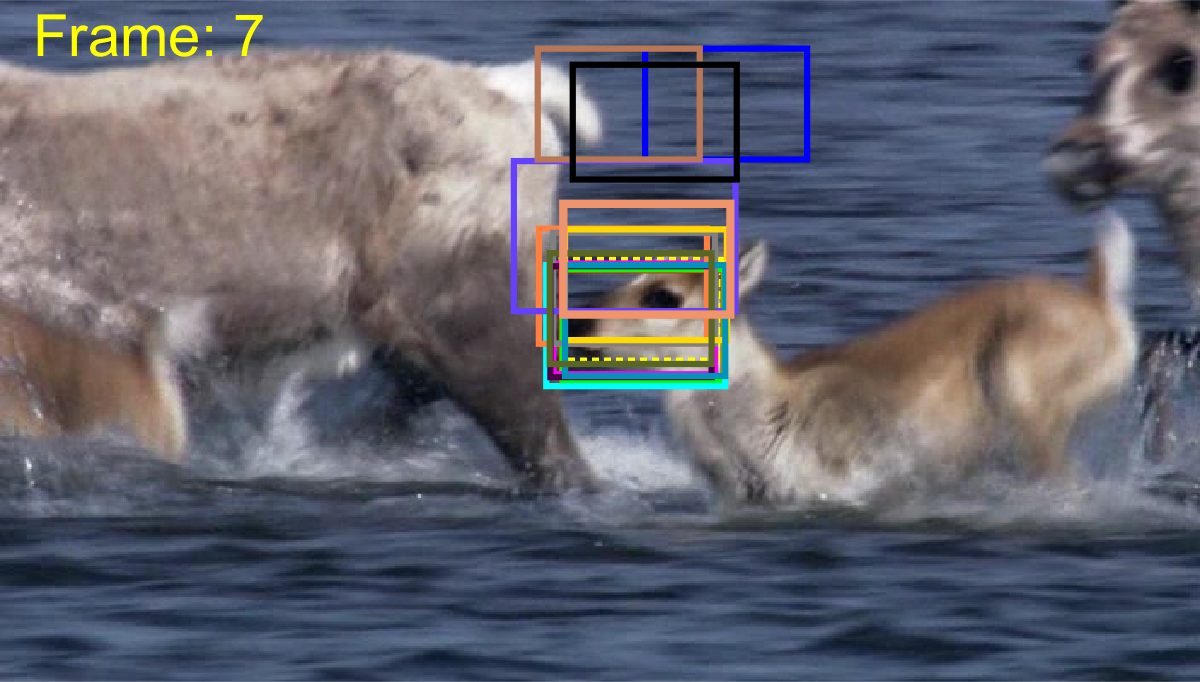}
\includegraphics[width= 0.32\linewidth]{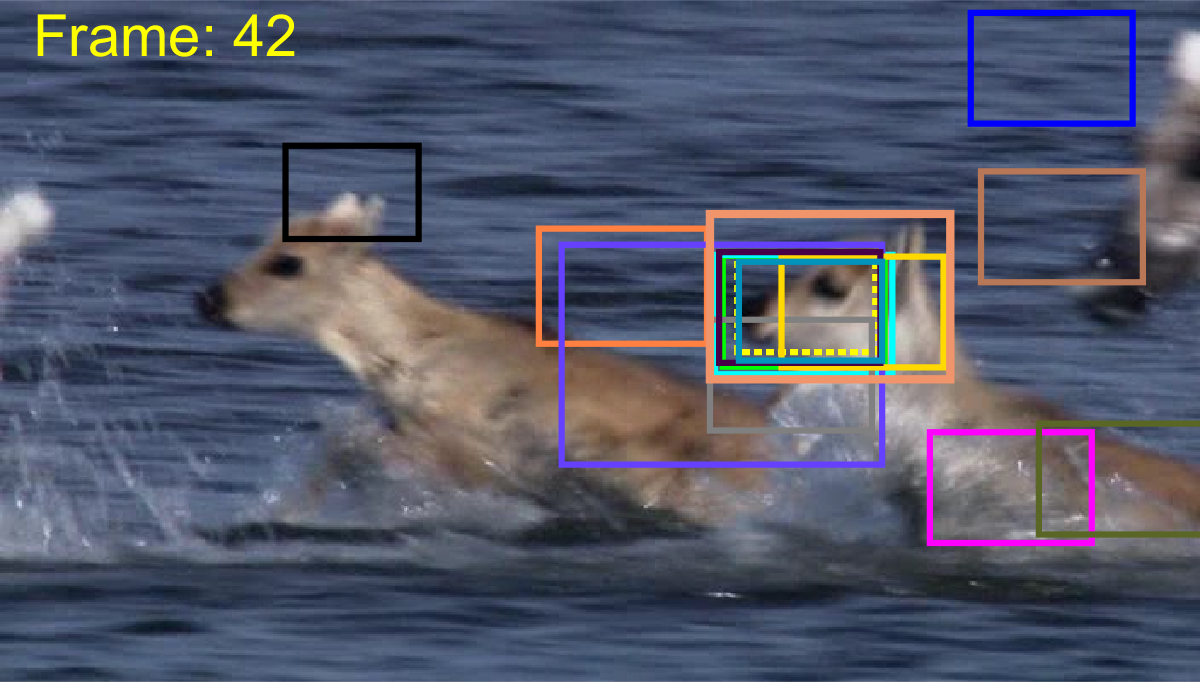}
\includegraphics[width= 0.32\linewidth]{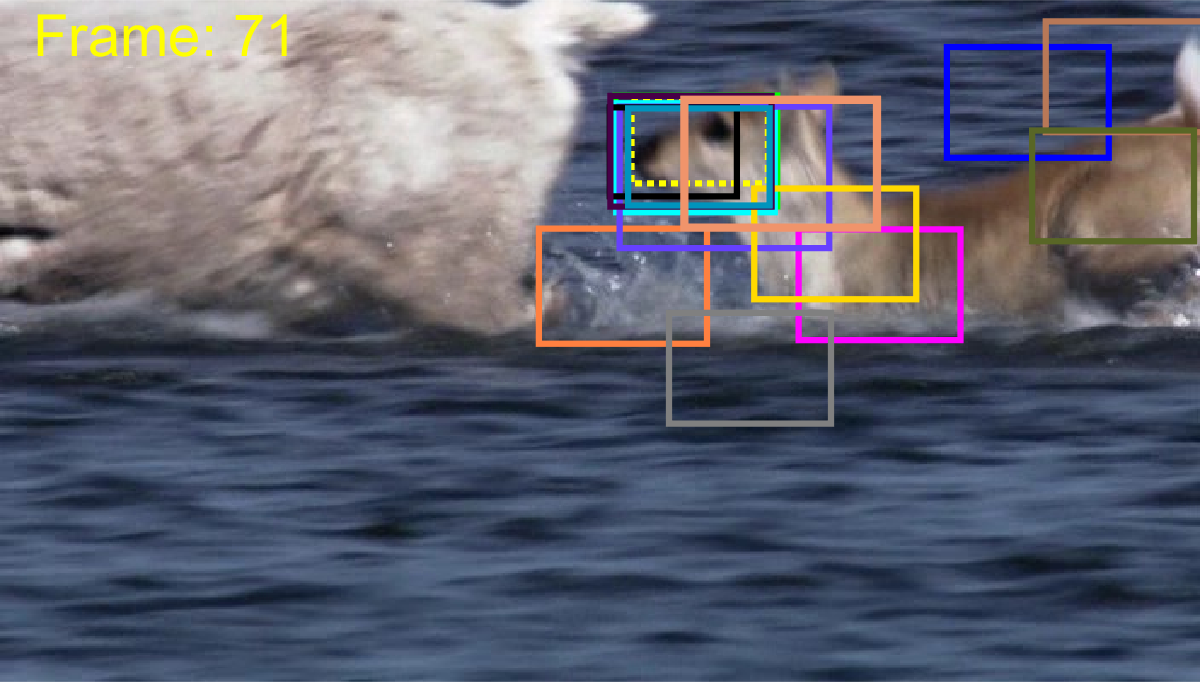}
\includegraphics[width= 0.32\linewidth]{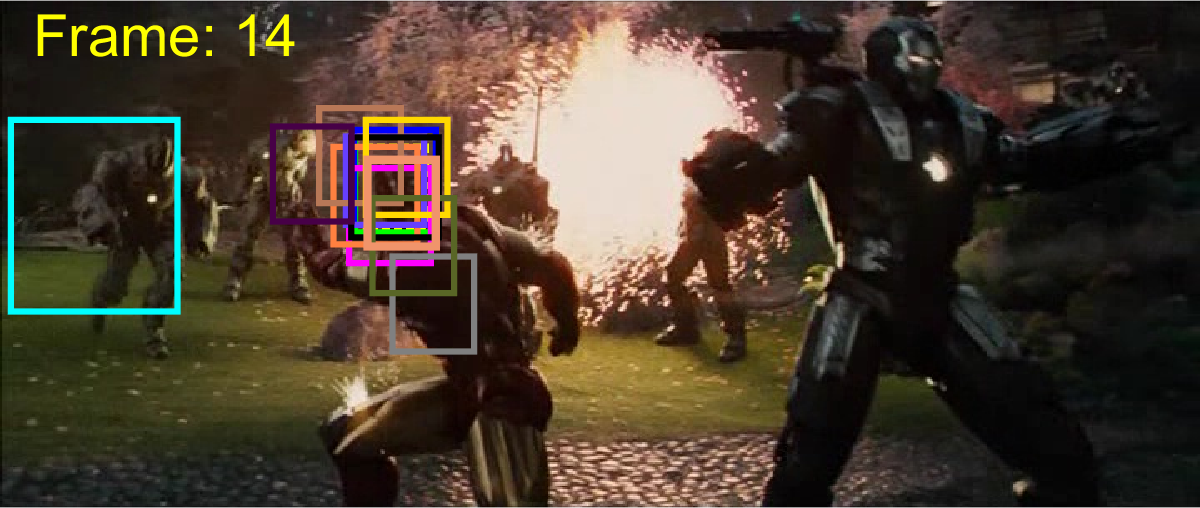}
\includegraphics[width= 0.32\linewidth]{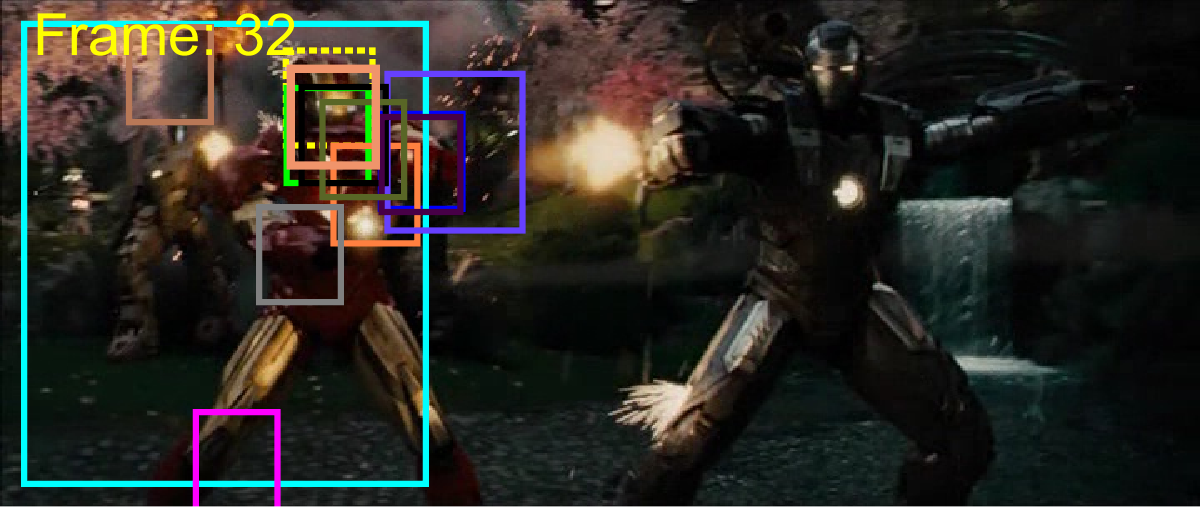}
\includegraphics[width= 0.32\linewidth]{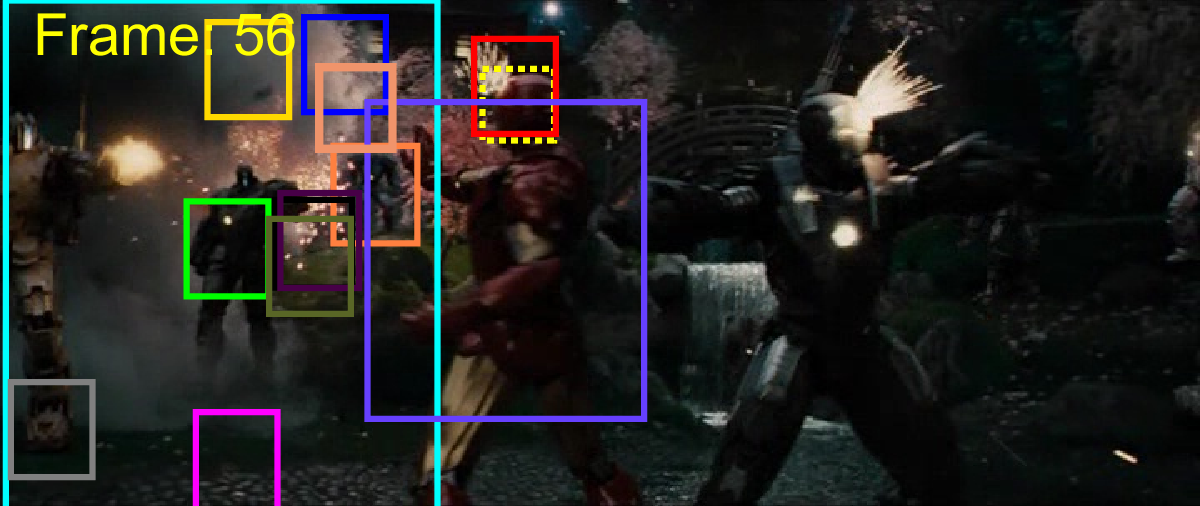}
\includegraphics[width= 0.32\linewidth]{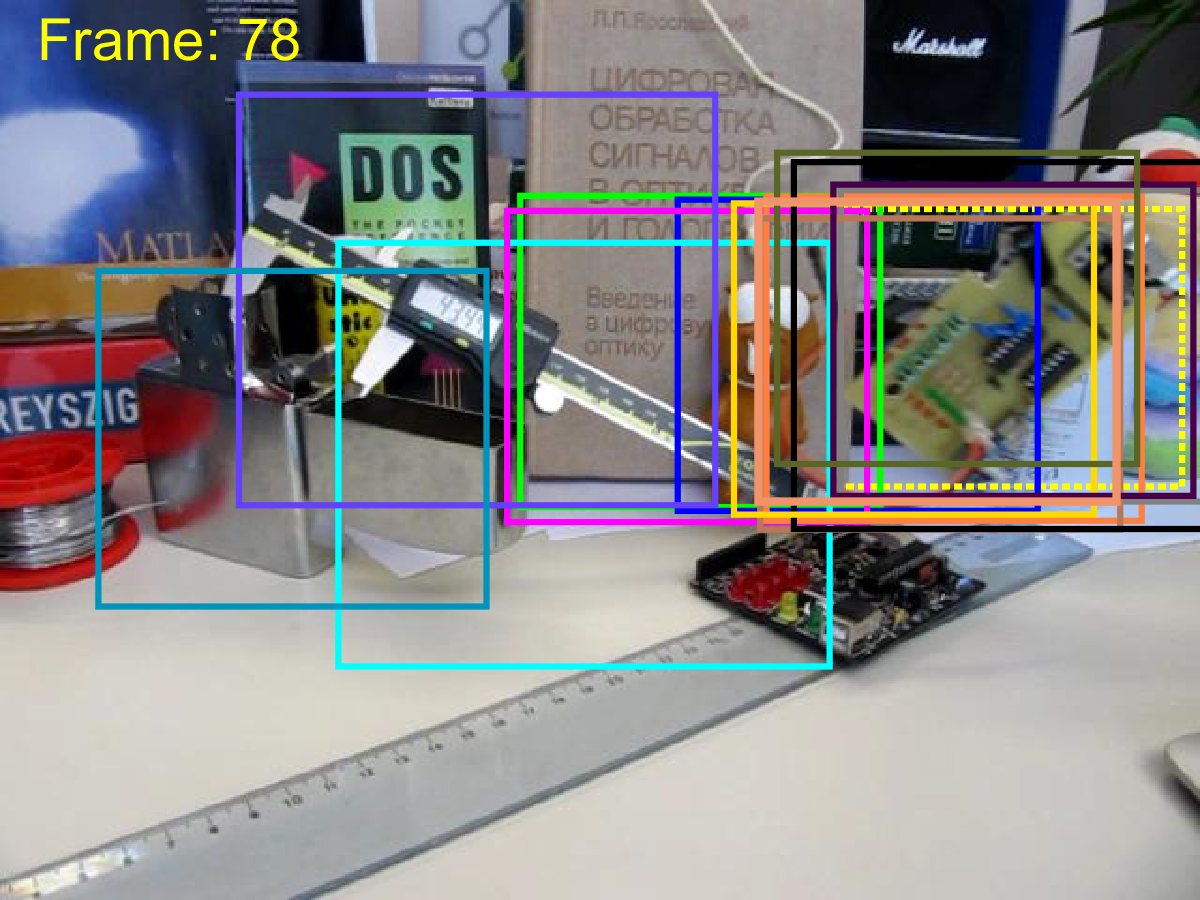}
\includegraphics[width= 0.32\linewidth]{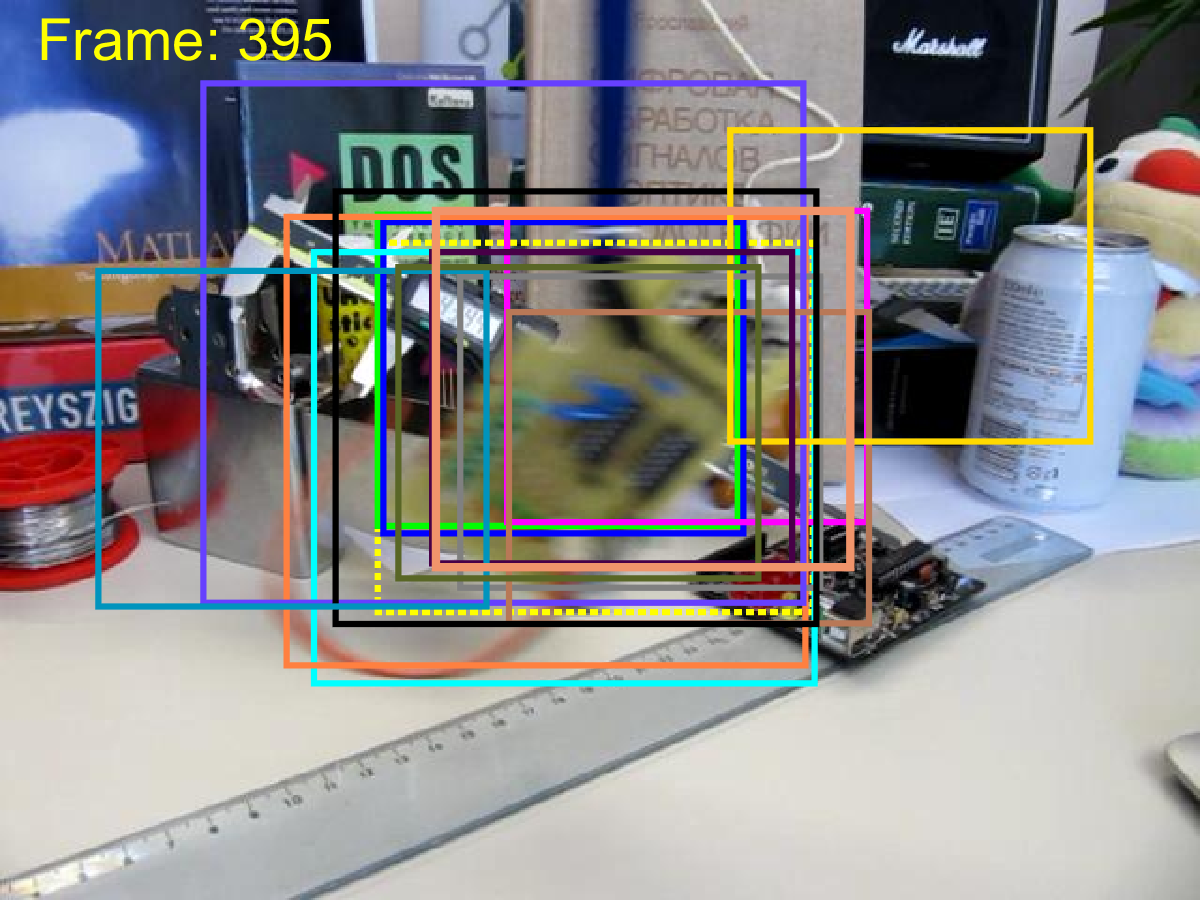}
\includegraphics[width= 0.32\linewidth]{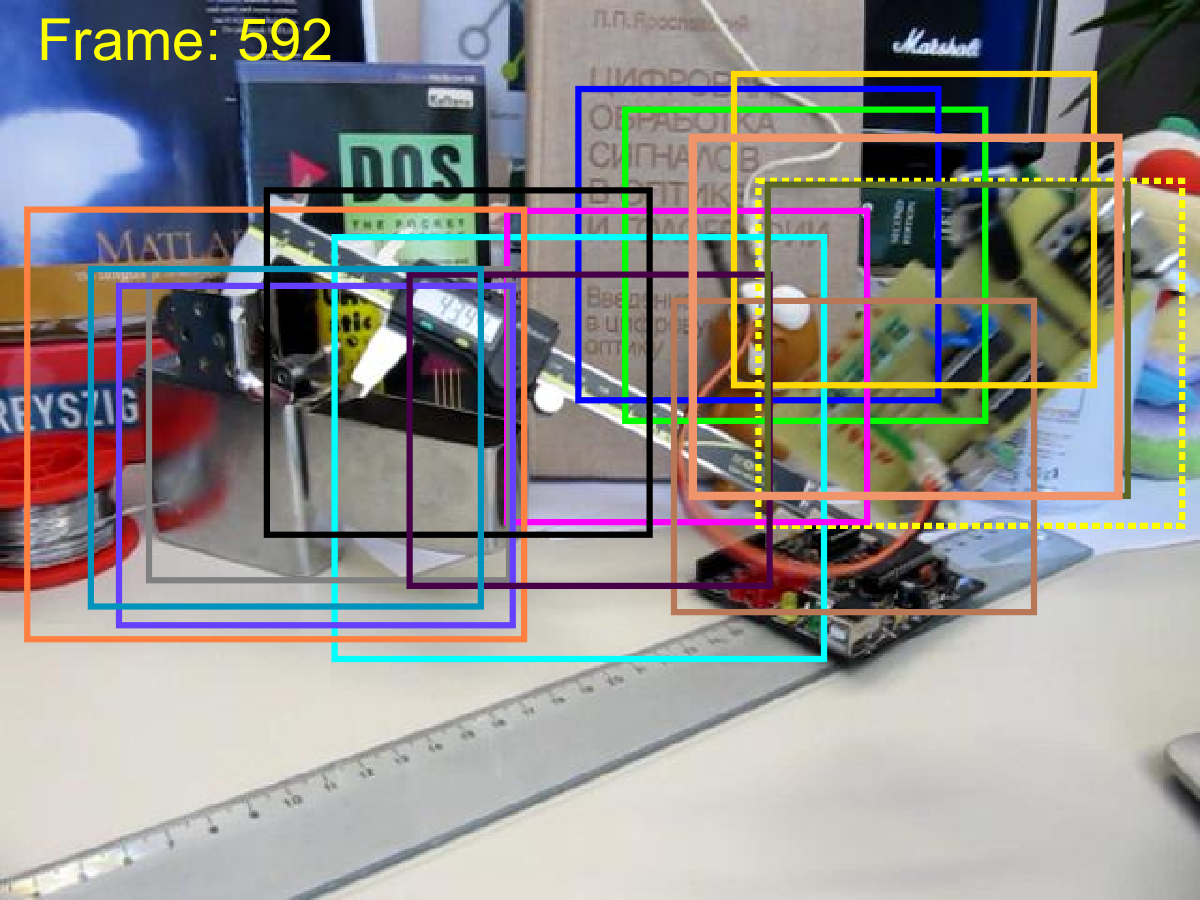}
\caption{Sample tracking results of evaluated algorithms on several challenging video sequences. \textbf{(from top)} \textit{FaceOcc2} with severe occlusions, \textit{Deer} with fast motion, \textit{Ironman} with in-plane and out-of-plane rotations, and \textit{Board} with background clutter.  In these sequences the orange box depicts the UTS, the yellow dashed line indicates the ground truth, and the rest depict the result of other trackers.}
\label{fig:eval_qual}
\vspace{-0.7 cm}
\end{figure}

Figure \ref{fig:eval_succ_challenges} depicts the performance of all of the investigated trackers. As it is evident from this plot, UST outperforms the other trackers by having the highest AUC. Interestingly, UST also outperform these trackers when facing illumination, scale and shape changes that show the resilience of this double appearance model used by the two detectors (Table \ref{tab:attributes}). Rotations, shear and fast motions are well addressed by the proposed tracker and only STRUCK handled motion blur as good as UST. However, background clutter and low resolution targets challenged the UST. Not equipped with the means to handle LR, the results seems acceptable. In the case of background clutter, the use of context information in CXT and local sparsity in LSK outperforms our proposed tracker, shedding some light on the future direction of research. Some examples of tracking results are depicted in Figure \ref{fig:eval_qual}.

\begin{table}[t]
\caption{Quantitative evaluation of five best trackers under different visual tracking challenges using AUC of success plot. The best performance for each attribute is \textbf{bold}}.
\label{tab:attributes}
\centering
\renewcommand{\arraystretch}{1.1}
\begin{tabularx}{\linewidth}{@{}l|X X X X X X@{}}
Attribute 					& CSK  & TLD  & LSK  & STRK & CXT  & UST \\ \hline
Illumination Variation     	& 0.40 & 0.49 & 0.50 & 0.46 & \textbf{0.52} & \textbf{0.52} \\
Deformation    				& 0.36 & 0.32 & 0.38 & 0.41 & 0.32 & \textbf{0.47} \\
Occlusion    				& 0.36 & 0.42 & 0.44 & 0.44 & 0.40 & \textbf{0.53} \\
Scale Variation    			& 0.34 & 0.44 & 0.46 & 0.43 & 0.45 & \textbf{0.51} \\
In-plane Rotation    		& 0.43 & 0.50 & 0.46 & 0.51 & 0.53 & \textbf{0.59} \\
Out-of-plane Rotation    	& 0.39 & 0.43 & 0.45 & 0.48 & 0.45 & \textbf{0.56} \\
Out-of-View     			& 0.32 & 0.45 & 0.39 & 0.44 & 0.38 & \textbf{0.52} \\
Low Resolution     			& 0.29 & 0.37 & \textbf{0.39} & \textbf{0.39} & 0.38 & 0.38 \\
Background Clutter     		& 0.42 & 0.40 & \textbf{0.45} & 0.39 & \textbf{0.45} & 0.41 \\
Fast Motion     			& 0.39 & 0.45 & 0.42 & 0.52 & 0.44 & \textbf{0.53} \\
Motion Blur     			& 0.32 & 0.42 & 0.37 & \textbf{0.48} & 0.38 & 0.46 \\
\hline
ALL    						& 0.41 & 0.46 & 0.46 & 0.48 & 0.48 & \textbf{0.52} \\
\end{tabularx}
\end{table}

Finally, it is prudent to note that UST achieved an average speed of 28.3 fps on a Pentium 4 Core i7 @ 3.2 GHz by Matlab/C++ implementation.
This experiment demonstrated that with a adequate information exchange in co-tracking, it is possible to balance a good trade-off between speed and accuracy, while the tracker is capable of properly under various tracking challenges.

\section{Conclusion}
\label{sect:conclusion}
The key component is a co-tracking framework consisting of a frequently updated (KNN-based) classifier and a more conservative (part-based) detector. Built upon uncertainty sampling foundation, samples deemed uncertain by the KNN classification are labeled by the part-based detector (the oracle). A memory budgeting mechanism keep classifier updates tractable. This accounting method, along with an optical-flow based ROI detection ensures that the proposed tracker, UST, meet th ereal-time criteria for tracking. Experimental results on challenging video sequences demonstrated that the UTS tracker achieve comparable accuracy to the state-of-the-art trackers, while outperforming them in terms of efficiency and robustness.


\bibliographystyle{IEEEtran}
\bibliography{ustbib}

\end{document}